\documentclass{article}

\usepackage{iclr2027_conference,times}

\usepackage{amsmath,amsfonts,bm}

\def\eqref#1{equation~\ref{#1}}

\def\1{\bm{1}}

\DeclareMathAlphabet{\mathsfit}{\encodingdefault}{\sfdefault}{m}{sl}
\SetMathAlphabet{\mathsfit}{bold}{\encodingdefault}{\sfdefault}{bx}{n}

\usepackage{amsmath}
\usepackage{amssymb}
\usepackage{booktabs}
\usepackage{graphicx}
\usepackage{multirow}
\usepackage[table]{xcolor}
\usepackage{hyperref}
\usepackage{url}
\hypersetup{hidelinks}

\newcommand{\method}{\textsc{VAMR}}
\newcommand{\qhpo}{\textsc{QHPO}}

\title{VAMR: Multi-Question Agentic Reasoning for Efficient Long-Form Video Understanding}

\newif\ifpublicversion
\publicversiontrue

\ifpublicversion
\iclrfinalcopy
\author{
\textbf{Runquan Gui}\textsuperscript{1}
\hspace{0.55em}
\textbf{Hanzhu Chen}\textsuperscript{1}
\hspace{0.55em}
\textbf{Zehao Wang}\textsuperscript{1}
\hspace{0.55em}
\textbf{Hanxin Zhu}\textsuperscript{1}
\hspace{0.55em}
\textbf{Xin Li}\textsuperscript{1}
\hspace{0.55em}
\textbf{Zhibo Chen}\textsuperscript{1}\thanks{Corresponding author:
\texttt{chenzhibo@ustc.edu.cn}. Contact: \texttt{runquangui@gmail.com}.}
\\[0.4ex]
{\normalfont\textsuperscript{1} University of Science and Technology of China}
}
\fi

\begin{document}

\maketitle
\ifpublicversion\lhead{Preprint. Under review.}\fi

\begin{abstract}
Long-form video understanding often involves multiple questions about different aspects of the same recording.
Yet existing video agents typically process each question through an isolated tool-use trajectory.
This repeatedly restarts video exploration and memory construction, missing opportunities to acquire evidence jointly and progressively build a shared understanding that supports the complete question set.
We introduce \textbf{VAMR} (\textbf{V}ideo \textbf{A}gent for \textbf{M}ulti-Question \textbf{R}easoning), which coordinates all questions about a video through one shared tool-use trajectory.
At each round, a persistent policy model can invoke tools for one or more unresolved questions and submit answers for questions with sufficient evidence.
Question-conditioned visual perception retrieves fine-grained clues for several questions in one call, while layered multi-question memory integrates reusable context into a shared video story and preserves separate evidence for individual questions.
After supervised fine-tuning initializes this interaction protocol, we propose question-horizon policy optimization (\qhpo) to optimize shared trajectories in which questions progress and finish at different rounds.
Specifically, a question-level critic estimates the value of each active question, while round alignment maps each question advantage to the rounds that directly serve it before the aligned advantages are aggregated to optimize the shared actor.
Across LVBench, Video-Holmes, and LongVideoBench, VAMR achieves the highest accuracy overall and the fewest reasoning rounds among iterative methods.
On LVBench, it reaches 62.1\% accuracy, exceeding VideoARM by \textbf{4.3} points while reducing reasoning rounds and processed frames by \textbf{85.9\%} and \textbf{61.4\%}.
\end{abstract}

\section{Introduction}
\label{sec:introduction}

Long-form video understanding has become a central problem for multimodal intelligence, requiring models to integrate evidence distributed across videos that may span hours~\citep{mangalam2023egoschema,fu2025videomme,wu2024longvideobench,wang2025lvbench}.
Both long-video benchmarks and practical systems often require multiple judgments about different aspects of the same recording: benchmark questions probe distinct events, dialogue, and temporal relationships~\citep{wang2025lvbench,chen2025cgbench}, while applications such as content moderation, security monitoring, autonomous driving, and content production must likewise assess several aspects of a common video or stream~\citep{aldahoul2024moderation}.
For example, a content moderation system may need to identify violent actions, harmful speech, and sensitive objects while also considering the context in which they appear.
A segment inspected for one question may also provide evidence for others, so processing the questions together can reuse visual observations and substantially reduce repeated exploration.
Efficiently answering multiple questions grounded in a common audiovisual context has therefore become an important and insufficiently addressed challenge for video understanding.

Recently, numerous approaches have emerged for long-form video reasoning.
End-to-end methods process sampled frames or compact video representations in a single pass, strengthening temporal reasoning within a provided visual context~\citep{feng2025videor1,park2025deepvideor1}.
To overcome the limited evidence available in this fixed context, iterative methods allow the model to select temporal regions and progressively refine its observations over multiple reasoning steps~\citep{he2025framethinker,pan2025timesearchr}.
More recent agentic methods extend this process by using a policy model to autonomously orchestrate perception tools, memory, and multi-step reasoning~\citep{zhang2025dvd,wang2026videohv,yin2025videoarm}.
Despite this progression from fixed-context reasoning to adaptive retrieval and autonomous tool use, existing methods remain centered on one question.
Answering multiple questions about the same video consequently repeats video exploration and memory construction, increasing computation while preventing questions from jointly guiding evidence acquisition and building a progressively enriched understanding of the video.

\begin{figure}[t]
\centering
\includegraphics[width=\textwidth]{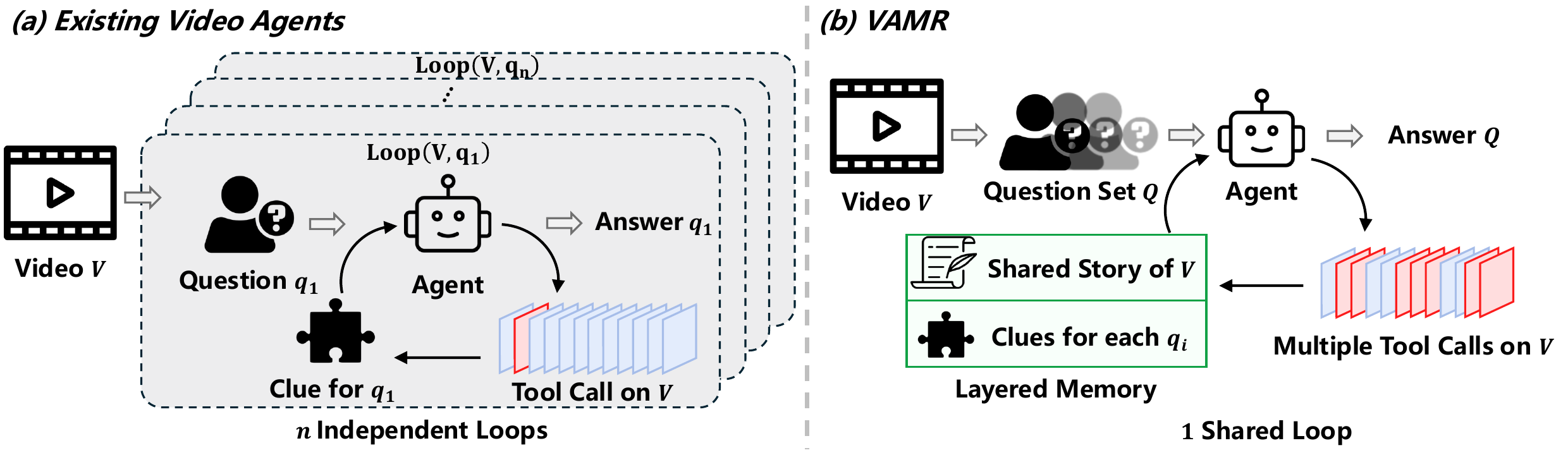}
\caption{Motivation for multi-question video reasoning. Existing video agents use separate online reasoning loops for the questions associated with a video, even when video-level preprocessing is reusable. VAMR instead coordinates the question set in one shared loop, allowing each tool call to enrich a shared video story and the separate clues maintained for individual questions.}
\label{fig:multi-question-motivation}
\end{figure}

To address this challenge, we introduce \method, a video agent that answers multiple questions through one coordinated reasoning process.
As illustrated in Figure~\ref{fig:multi-question-motivation}, VAMR formulates agentic multi-question inference as a shared sequential decision process, replacing independent per-question loops with one ReAct loop over the complete question set~\citep{yao2023react}.
At each step, the policy decides which questions to answer and selects segments and unresolved questions for evidence retrieval.
The QA-inspector then retrieves fine-grained clues for the selected questions, while layered memory integrates reusable context into a shared video story and preserves separate evidence for individual questions.
After supervised fine-tuning (SFT) initializes this interaction protocol, question-horizon policy optimization (\qhpo{}) learns a separate value for each active question instead of assigning one value to the shared state.
It then retains each question-level advantage only at rounds containing a validated action for that question and aggregates the aligned advantages to train the round-level actor over joint actions.
Together, the shared formulation and QHPO allow questions to progress and terminate at different rounds while building a persistent understanding of the video.

Experiments on LVBench~\citep{wang2025lvbench}, Video-Holmes~\citep{cheng2025videoholmes}, and LongVideoBench~\citep{wu2024longvideobench} evaluate both reasoning accuracy and efficiency.
Under our unified evaluation setting, VAMR achieves the highest accuracy on all three benchmarks while requiring the fewest reasoning rounds.
On LVBench, VAMR reaches 62.1\% accuracy, exceeding VideoARM by 4.3 points while reducing reasoning rounds and processed frames by 85.9\% and 61.4\%.
The gains remain consistent on Video-Holmes and LongVideoBench, which represent short-video and low-question-density settings, respectively.
Architecture ablations examine how the perception and memory design supports the shared loop, while training ablations isolate the contribution of QHPO.
Controlled analyses show that the benefit grows as more questions are coordinated within one loop and remains substantial across all relative evidence-overlap quartiles.

Our contributions are:
\begin{itemize}
\item We introduce VAMR, which replaces independent per-question loops with coordinated reasoning over a shared video understanding.
To our knowledge, VAMR is the first video agent to coordinate reasoning across multiple questions within a shared trajectory.
\item We propose \qhpo{}, a policy optimization algorithm for shared multi-question trajectories in which questions progress and terminate at different rounds.
QHPO uses a question-level critic to estimate per-question advantages, aligns them to the rounds that serve each question, and aggregates them to optimize a round-level actor over joint actions.
\item Experiments on three long-video benchmarks demonstrate consistent gains in accuracy and reasoning efficiency.
Ablations and analyses identify the contributions of the shared architecture and QHPO, showing robust gains across question densities and evidence overlap.
\end{itemize}

\section{Related Work}
\label{sec:related}

\paragraph{Long-form video reasoning.}
Long-video reasoning depends not only on a model's reasoning ability, but also on whether relevant evidence enters its limited visual context.
End-to-end methods such as Video-R1~\citep{feng2025videor1} and DeepVideo-R1~\citep{park2025deepvideor1} strengthen temporal reasoning over a fixed visual context.
Iterative methods such as FrameThinker~\citep{he2025framethinker} and TimeSearch-R~\citep{pan2025timesearchr} instead adapt that context through hierarchical representations or temporal search~\citep{wang2024videotree,ye2025tstar,wang2025videorts}.
More recently, agentic methods place an LLM in control of tool use and memory for iterative evidence acquisition~\citep{zhang2025dvd,wang2026videohv,yin2025videoarm,yan2026ava}.
Deep Video Discovery combines global browsing, semantic clip search, and frame-level inspection over a preconstructed video database~\citep{zhang2025dvd}.
VideoARM further replaces preconstructed databases with online hierarchical memory and invokes coarse-to-fine perception tools on demand~\citep{yin2025videoarm}.
Both representative approaches organize online policy state, adaptive retrieval, and stopping around individual queries.
\method{} instead maintains one persistent loop that jointly schedules question-specific evidence needs and propagates each observation through shared and question-specific memory.

\paragraph{Cross-question information use.}
Prior work exploits semantic relations among questions about the same video as auxiliary information for one-pass answer prediction or feature extraction~\citep{lei2020multiquestion,pei2025collective,liang2025isolated}.
UniVA packages multiple interdependent questions into a single composite objective and returns all answers in one inference through a general Plan-and-Act workflow~\citep{liang2025univa}.
VAMR instead maintains the questions as distinct decision objectives throughout inference, allowing them to jointly direct tool use, retain separate evidence, and terminate asynchronously.
To our knowledge, VAMR is the first video agent to coordinate reasoning across multiple questions within a shared trajectory.

\paragraph{Policy optimization for tool-using agents.}
Policy optimization for tool-using agents must assign outcome credit across reasoning turns, tool calls, and environment feedback.
Existing approaches either learn intermediate values or rewards for multi-turn interactions~\citep{zhou2024archer,wang2025stepsearch} or derive localized advantages from group-relative outcome signals~\citep{jin2025searchr1,li2025torl,wang2025ragen,feng2025gigpo}.
Related video-agent training methods optimize adaptive perception or tool use for a single task~\citep{zhang2026eva,zeng2026videoo3}.
These methods generally associate a tool-use trajectory with one task outcome and therefore do not address joint actions that advance multiple questions with different termination times.
\qhpo{} instead estimates question-level values, aligns them with directly served rounds, and aggregates the retained advantages for round-level updates within a shared multi-question trajectory.

\section{VAMR: Video Agent for Multi-Question Reasoning}
\label{sec:method}

This section formulates multi-question reasoning for tool-using video agents as a shared sequential decision problem and presents \method{} as its concrete realization.
We first contrast independent and shared tool-use loops, then describe how VAMR implements the shared process through a bounded active question window, question-conditioned perception, and layered memory.

\subsection{Multi-question video reasoning}

Let $V$ be a video and $Q=\{q_i\}_{i=1}^{n}$ its associated multiple-choice questions, where $q_i$ is paired with a set of answer options $\mathcal{A}_i$.
The ground-truth and predicted answers are denoted by $y_i,\hat y_i\in\mathcal{A}_i$, respectively, and $t$ indexes interaction turns.

\paragraph{Independent loops.}
Most video agents are organized around a single query.
For question $q_i$, a policy $\pi_\theta$ with parameters $\theta$ conditions each decision on that question and a private running memory $m_t^i$.
At each turn, the policy selects one of three action types:
\begin{equation}
\begin{aligned}
a_t^i
&\sim\pi_\theta(\cdot\mid q_i,m_t^i),\\
a_t^i
&\in\{\operatorname{answer}(\hat y_i),
\operatorname{visual\text{-}inspector}(V_{[l,r]}),
\operatorname{ASR}(V_{[l,r]})\}.
\end{aligned}
\label{eq:single-question-loop}
\end{equation}
Here, $V_{[l,r]}$ denotes the video segment within the requested temporal interval $[l,r]$.
The visual inspector summarizes sampled frames from this segment, while ASR transcribes its speech.
The resulting observation is incorporated into the private memory $m_{t+1}^i$ before the next decision.
An $\operatorname{answer}$ action terminates the current trajectory.
A video with $n$ questions is therefore processed through $n$ independent trajectories, each with its own observations and memory.

\paragraph{Shared loop.}
Independent loops cannot coordinate adaptive retrieval or carry their online reasoning state across questions.
To enable cross-question coordination, we propose a shared-loop formulation that models all questions about the same video as concurrent objectives within one sequential decision process.
Specifically, we maintain one policy model and one memory $M_t$ across the complete question set.
Let $B_t\subseteq Q$ be the active question window exposed to the policy at turn $t$, with the remaining questions kept in an ordered backlog.
We formulate each policy output as a joint action that may contain answer submissions and tool calls for arbitrary active questions:
\begin{equation}
\begin{aligned}
\mathbf{a}_t
&\sim\pi_\theta(\cdot\mid B_t,M_t),\\
\mathbf{a}_t
&\subseteq\{\operatorname{answer}(\hat y_i),
\operatorname{visual\text{-}inspector}(V_{[l,r]}),
\operatorname{ASR}(V_{[l,r]})\}_{q_i\in B_t}.
\end{aligned}
\label{eq:multi-question-loop}
\end{equation}
Unlike the scalar action $a_t^i$ in an independent trajectory, $\mathbf{a}_t$ may include multiple answer submissions and tool calls in one turn.
After execution, tool observations update $M_{t+1}$, while answered questions leave the active window and are replaced from the backlog.
The shared loop remains active until all questions are answered.
Our formulation makes cross-question coordination part of the agent trajectory rather than a batching rule: the policy decides each action jointly over active questions, retrieved observations accumulate in a persistent state that informs subsequent decisions, and questions terminate asynchronously.
The task objective remains average accuracy across questions.

\subsection{VAMR framework}

\begin{figure}[t]
\centering
\includegraphics[width=\textwidth]{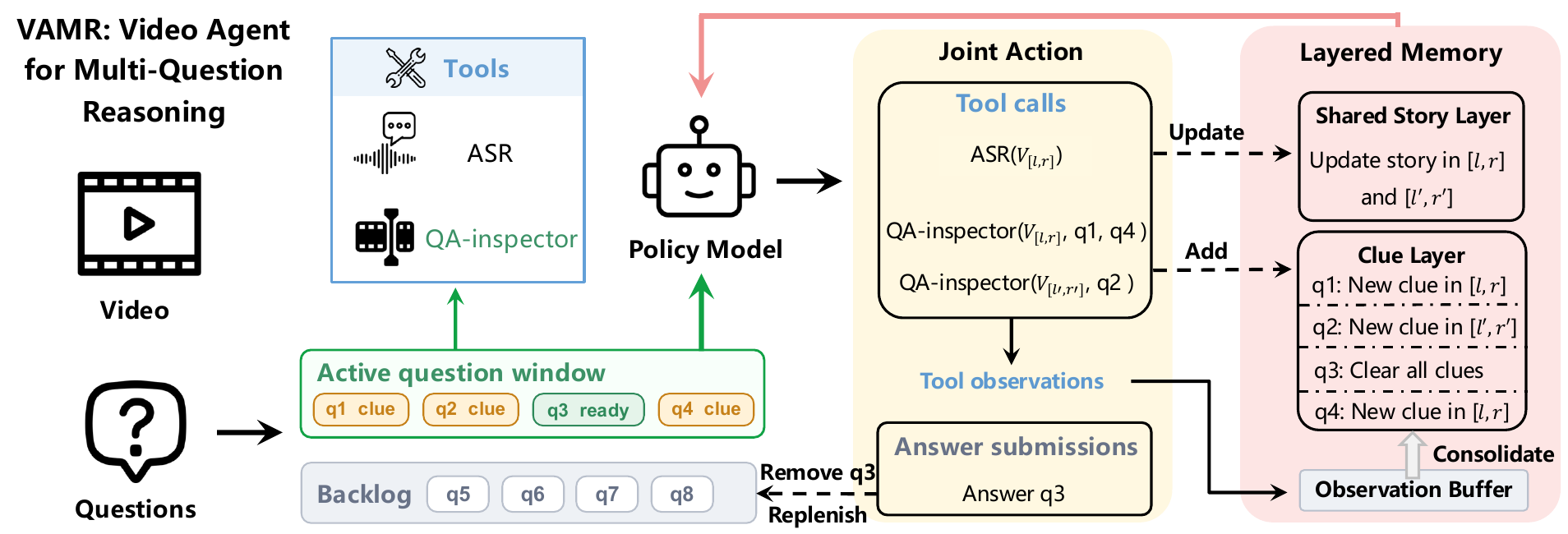}
\caption{Overview of VAMR. One persistent policy model coordinates the active question window, submits resolved answers, and invokes ASR or question-conditioned visual tools. Tool observations update the shared story layer, question clue layer, and observation buffer, while completed slots are replenished from the backlog.}
\label{fig:vamr-overview}
\end{figure}

Figure~\ref{fig:vamr-overview} illustrates how VAMR instantiates the shared loop in Eq.~\ref{eq:multi-question-loop}.
A persistent policy operates over the active window $B_t$ and memory $M_t$, deciding which questions to answer and which segments and question-option pairs to retrieve.
To support coordinated decisions throughout this trajectory, VAMR introduces question-conditioned perception and layered memory, which build reusable video understanding from local retrievals while organizing evidence for individual questions.
The QA-inspector retrieves reusable context and separate clues from each policy-selected segment, while memory consolidates them into a shared story and question-specific clue states for subsequent decisions.
Detailed execution rules are provided in Appendix~\ref{app:agentic-framework-details}.

\paragraph{Question-conditioned shared perception.}
The proposed QA-inspector extends a generic visual inspector with two complementary capabilities: question-targeted clue retrieval and shared retrieval for multiple questions.
Given a temporal interval and target set $Q_t^{\mathrm{tar}}$ selected by the policy, the tool receives the corresponding questions and answer options:
\begin{equation}
\left(g_t,\{e_t^i\}_{q_i\in Q_t^{\mathrm{tar}}}\right)
=
\operatorname{QA\text{-}inspector}
\left(V_{[l,r]},
\{(q_i,\mathcal{A}_i)\}_{q_i\in Q_t^{\mathrm{tar}}}\right).
\label{eq:qa-inspector}
\end{equation}
The shared output $g_t$ summarizes the observed content, while $e_t^i$ records the evidence relevant to question $q_i$.
First, conditioning retrieval on question-option pairs shifts the tool from generic clip description to fine-grained clues that discriminate among candidate answers.
Second, one retrieval call can serve several policy-selected questions, returning a reusable account of the segment together with distinct clues for each question.
This design enables cross-question reuse during retrieval without collapsing the different evidence requirements of individual questions.

\paragraph{Layered multi-question memory.}
A single running summary is not sufficient when questions require different evidence.
VAMR maintains a shared story layer alongside a question clue layer that contains an independent clue memory for every active question:
\begin{equation}
M_t=\left(G_t,\{E_t^i\}_{q_i\in B_t},T_t\right).
\label{eq:vamr-memory}
\end{equation}
The shared story layer $G_t$ maintains a reusable time-anchored account of the video, while each clue memory $E_t^i$ retains the evidence and unresolved information specific to question $q_i$.
The observation buffer $T_t$ temporarily holds recent outputs $g_t$ and $\{e_t^i\}_{q_i\in Q_t^{\mathrm{tar}}}$ before consolidation.
When an answer is accepted, VAMR integrates reusable buffered content into $G_t$, routes question-specific clues into the corresponding $E_t^i$, discards the clue memories of completed questions, and clears the buffer before activating new questions.
Each new question therefore inherits the accumulated shared story but begins with its own clue memory, preserving video-level knowledge across question transitions without carrying over irrelevant question-specific evidence.

\section{Learning the VAMR policy}
\label{sec:training}

In this section, we describe the two-stage procedure for learning the VAMR policy to operate within the shared multi-question loop.
We first use supervised fine-tuning on multi-question trajectories to initialize the joint action protocol, and then apply QHPO to improve retrieval and stopping through question-conditioned value estimation and direct round alignment.

\subsection{Two-stage SFT initialization}
\label{sec:sft}

Each SFT example is one policy-model turn from an executed \method{} trajectory.
Following Section~\ref{sec:method}, let $B_t$ denote the active questions and $M_t$ the layered memory.
For SFT construction only, $P_t=\{(y_i,\mathcal{I}_i)\mid q_i\in B_t\}$ contains the correct answer $y_i$ and annotated evidence intervals $\mathcal{I}_i$ for each active question.
We construct each example in two stages:
\begin{equation}
\begin{aligned}
(B_t,M_t,P_t)&\longrightarrow \mathbf{a}_t^\star
&&\text{Stage 1: action selection},\\
(B_t,M_t,\mathbf{a}_t^\star)&\longrightarrow z_t^\star
&&\text{Stage 2: reasoning generation}.
\end{aligned}
\label{eq:two-stage-sft}
\end{equation}
Stage 1 uses $P_t$ to select a joint action $\mathbf{a}_t^\star$ containing supported answer submissions and tool calls for unresolved questions, which is then executed to advance the trajectory.
After the action is fixed, Stage 2 removes $P_t$ and generates its reasoning trace $z_t^\star$ from $(B_t,M_t,\mathbf{a}_t^\star)$.
The resulting example maps $(B_t,M_t)$ to $(z_t^\star,\mathbf{a}_t^\star)$, so privileged information guides action selection without entering the policy input or reasoning trace.
Standard autoregressive training on these examples initializes the joint action protocol for QHPO.
Construction details are provided in Appendix~\ref{app:sft-data-construction}.

\subsection{Question-horizon policy optimization}
\label{sec:qhpo}

After SFT initializes the joint action protocol, we propose QHPO to further optimize retrieval and stopping decisions using complete shared-loop trajectories sampled from the current policy.
At round $t$, the policy observes $(B_t,M_t)$ and produces reasoning $z_t$ together with joint action $\mathbf{a}_t$.
After the trajectory terminates, each question $q_i$ receives reward $R_i=1$ if the policy submits the correct answer, and $R_i=0$ otherwise.
Since questions within the same shared state may have different probabilities of success, a single round-level value cannot distinguish their learning signals.
QHPO instead introduces a question-level critic $V_\phi(B_t,M_t,q_i)\in(0,1)$, parameterized by $\phi$, to estimate $\mathbb{E}_{\pi_\theta}[R_i\mid B_t,M_t,q_i]$.
The critic is trained by
\begin{equation}
\mathcal{L}_{\mathrm{critic}}
=
-\mathbb{E}_{i,t:q_i\in B_t}\left[
R_i\log V_\phi(B_t,M_t,q_i)
+(1-R_i)\log\left(1-V_\phi(B_t,M_t,q_i)\right)
\right].
\label{eq:qhpo-critic}
\end{equation}
Each active question receives a separate value estimate within the shared policy state.

QHPO next applies generalized advantage estimation~\citep{schulman2016gae} separately to each question, using its critic values and terminal reward to obtain a question-level advantage $A_{t,i}$ for every round in which $q_i$ is active.
For each question, we identify the rounds whose joint actions directly serve that question, such as a targeted visual call or its answer submission.
Only these aligned rounds retain the corresponding question-level advantage, preventing an outcome from updating rounds that contain no direct action for that question.
Let $\mathcal{S}_t=\{i:q_i\in B_t,\ A_{t,i}\ne0\}$ denote the indices of questions with a nonzero advantage at round $t$.
The round-level advantage is the average of their question-level advantages:
\begin{equation}
A_t
=
\frac{1}{|\mathcal{S}_t|}
\sum_{i\in\mathcal{S}_t}A_{t,i}.
\label{eq:question-horizon-credit}
\end{equation}
Because the policy emits each joint output as one sequence, this aggregation retains a single learning signal for the complete joint action rather than assigning separate objectives to its components.

Once $A_t$ has been obtained, actor optimization follows standard clipped PPO~\citep{schulman2017proximal}.
Let $\rho_t(\theta)$ denote the importance ratio.
The actor objective is
\begin{equation}
\mathcal{L}_{\mathrm{actor}}
=
-\mathbb{E}_t\left[
\min\left(
\rho_t(\theta)A_t,
\operatorname{clip}
\left(\rho_t(\theta),1-\epsilon,1+\epsilon\right)A_t
\right)
\right].
\label{eq:qhpo-actor}
\end{equation}
Here, $\epsilon$ is the PPO clipping threshold.
After critic initialization, the actor and critic are updated jointly.
In summary, QHPO estimates outcomes at the question level, conservatively aligns each advantage with directly served rounds, and then optimizes the shared actor with the aggregated round-level signal.
Further optimization and credit-assignment details are provided in Appendix~\ref{app:training-details}.

\section{Experiments}
\label{sec:experiments}

\subsection{Experimental Setup}

\paragraph{Datasets.}
We train VAMR exclusively on CG-Bench~\citep{chen2025cgbench}, partitioned by video identifier.
The SFT and \qhpo{} partitions contain 5,775 and 6,001 question instances, respectively, with a disjoint subset held out for validation.
We evaluate on LVBench~\citep{wang2025lvbench}, Video-Holmes~\citep{cheng2025videoholmes}, and LongVideoBench~\citep{wu2024longvideobench}.
LVBench has the highest question density, averaging 15.04 questions per video compared with 6.80 on Video-Holmes and 1.78 on LongVideoBench.
It also contains the longest videos, averaging 68.4 minutes compared with 7.9 minutes for LongVideoBench and 2.8 minutes for Video-Holmes.
With both the longest videos and the most questions per video, LVBench represents the most challenging benchmark in our evaluation suite.
Detailed statistics are provided in Appendix~\ref{app:datasets}.

\paragraph{Models and baselines.}
VAMR uses Qwen3.5-9B~\citep{qwen2026qwen35} as the policy model, Qwen3.7-Plus~\citep{alibaba2026qwen37plus} for visual perception, and Qwen3-ASR-Flash~\citep{alibaba2026qwen3asr} for speech transcription.
The controls include direct answering with Qwen3.5-9B or Qwen3.7-Plus, an untrained Qwen3.5-9B tool-use loop, and a separately trained independent-loop policy with matched supervision and training budget.
We compare three baseline categories.
\textit{(1) End-to-end video reasoning} includes Video-R1~\citep{feng2025videor1} and DeepVideo-R1~\citep{park2025deepvideor1}.
\textit{(2) Iterative video reasoning} includes FrameThinker~\citep{he2025framethinker} and TimeSearch-R~\citep{pan2025timesearchr}.
\textit{(3) Agentic video reasoning} includes DVD~\citep{zhang2025dvd}, VideoHV-Agent~\citep{wang2026videohv}, and VideoARM~\citep{yin2025videoarm}.
All reproduced methods are evaluated on the same benchmark examples under common answer-scoring and efficiency-accounting rules.
Additional results under native model configurations are provided in Appendix~\ref{app:main-experiment-details}.

\paragraph{Metrics.}
We report accuracy and two complementary efficiency measures.
Rounds/Q normalizes the number of reasoning rounds by the number of questions and captures the sequential interaction required by an agent.
Frames/Q measures the visual workload processed by perception models.
Every frame is counted whenever it is processed, including repeated inputs across rounds.
Together, Rounds/Q and Frames/Q provide complementary measures of reasoning efficiency by capturing sequential interaction and visual processing.
Detailed accounting rules are provided in Appendix~\ref{app:main-experiment-details}.

\subsection{Main Results}

As shown in Table~\ref{tab:main-results}, VAMR achieves the highest accuracy with the fewest reasoning rounds across all three benchmarks under our unified evaluation setting.
These consistent improvements in both effectiveness and efficiency across different video durations and question densities demonstrate strong generalization.
This advantage is especially clear on LVBench.
VAMR requires only 0.9 Rounds/Q and 84.9 Frames/Q, compared with 6.4 Rounds/Q and 220.2 Frames/Q for VideoARM, reducing reasoning rounds by \textbf{85.9\%} and processed frames by \textbf{61.4\%}.
Despite using substantially less perception and interaction, VAMR reaches 62.1 accuracy and surpasses VideoARM by \textbf{+4.3} points.
These simultaneous gains show that VAMR improves evidence-acquisition efficiency while building a stronger shared video understanding, rather than trading accuracy for efficiency.
With matched backbone, tools, supervision, and training budget, VAMR improves over its separately trained independent-loop counterpart by \textbf{5.7} accuracy points while reducing Rounds/Q from 4.1 to 0.9 and Frames/Q from 224.0 to 84.9.
The concurrent improvements in accuracy and efficiency directly demonstrate the advantage of the shared loop over solving each question independently.

\begin{table}[t]
\caption{Main results on three held-out video benchmarks. The best and second-best results are shown in bold and underlined, respectively.}
\label{tab:main-results}
\begin{center}
\setlength{\tabcolsep}{4.5pt}
\renewcommand{\arraystretch}{1.4}
\resizebox{\textwidth}{!}{%
\begin{tabular}{lccccccccc}
\toprule
\multirow{2}{*}{\textbf{Method}}
& \multicolumn{3}{c}{\textbf{LVBench}}
& \multicolumn{3}{c}{\textbf{Video-Holmes}}
& \multicolumn{3}{c}{\textbf{LongVideoBench}} \\
\cmidrule(lr){2-4} \cmidrule(lr){5-7} \cmidrule(lr){8-10}
& \textbf{Acc. $\uparrow$}
& \textbf{Rounds/Q $\downarrow$}
& \textbf{Frames/Q $\downarrow$}
& \textbf{Acc. $\uparrow$}
& \textbf{Rounds/Q $\downarrow$}
& \textbf{Frames/Q $\downarrow$}
& \textbf{Acc. $\uparrow$}
& \textbf{Rounds/Q $\downarrow$}
& \textbf{Frames/Q $\downarrow$} \\
\midrule

\rowcolor{gray!15}
\multicolumn{10}{c}{\textit{Backbone and Loop Controls}} \\
Qwen3.5-9B (Direct)
& 28.8 & N/A & \textbf{32.0}
& 43.8 & N/A & \underline{32.0}
& 58.9 & N/A & \underline{32.0} \\
Qwen3.7-Plus (Direct)
& 43.7 & N/A & \textbf{32.0}
& 49.8 & N/A & \underline{32.0}
& 62.4 & N/A & \underline{32.0} \\
Qwen3.5-9B + Tools
& 45.2 & 4.8 & 210.7
& 46.1 & 3.4 & 126.7
& 60.3 & 4.5 & 161.0 \\
Ind. Loops (SFT+PPO)
& 56.4 & 4.1 & 224.0
& 53.8 & 3.1 & 138.6
& 65.1 & 4.2 & 192.4 \\

\midrule
\rowcolor{cyan!10}
\multicolumn{10}{c}{\textit{End-to-End Video Reasoning}} \\
Video-R1
& 35.6 & N/A & \textbf{32.0}
& 38.5 & N/A & \underline{32.0}
& 58.3 & N/A & \underline{32.0} \\
DeepVideo-R1
& 37.1 & N/A & \textbf{32.0}
& 39.3 & N/A & \underline{32.0}
& 57.5 & N/A & \underline{32.0} \\

\midrule
\rowcolor{cyan!10}
\multicolumn{10}{c}{\textit{Iterative Video Reasoning}} \\
FrameThinker
& 36.7 & 4.0 & \underline{50.0}
& 45.0 & \underline{2.5} & \textbf{22.4}
& 50.9 & \underline{2.4} & \textbf{28.8} \\
TimeSearch-R
& 43.5 & \underline{3.6} & 2779.6
& 36.3 & 3.6 & 1563.2
& 60.6 & 3.7 & 2882.2 \\

\midrule
\rowcolor{cyan!10}
\multicolumn{10}{c}{\textit{Agentic Video Reasoning}} \\
DVD
& 53.1 & 7.3 & 881.6
& 50.4 & 6.3 & 317.3
& 65.2 & 7.5 & 862.4 \\
VideoHV-Agent
& 54.6 & 5.7 & 287.0
& 52.3 & 5.3 & 35.6
& 61.8 & 5.4 & 277.0 \\
VideoARM
& \underline{57.8} & 6.4 & 220.2
& \underline{54.6} & 4.4 & 135.3
& \underline{65.9} & 6.6 & 185.9 \\

\midrule
\rowcolor{cyan!10}
\multicolumn{10}{c}{\textbf{Our Method}} \\
\textbf{\method{}} (Ours)
& \textbf{62.1} & \textbf{0.9} & 84.9
& \textbf{58.2} & \textbf{1.4} & 98.4
& \textbf{66.5} & \textbf{1.8} & 175.0 \\

\bottomrule
\end{tabular}
}
\end{center}
\end{table}

The benefits of VAMR are not confined to dense groups of questions over long videos.
On Video-Holmes, whose videos average only 2.8 minutes, VAMR improves over the trained independent-loop counterpart by 4.4 points while reducing reasoning rounds and processed frames by 54.8\% and 29.0\%.
Although short videos leave less redundant temporal search to eliminate, the substantial accuracy gain confirms that shared multi-question reasoning remains effective in this setting.
On LongVideoBench, with only 1.78 questions per video, the accuracy margin narrows to 1.4 points, while VAMR still reduces reasoning rounds by 57.1\% and processed frames by 9.0\%.
Although lower question density limits the available accuracy and visual-efficiency gains, VAMR remains effective and retains a sequential-round advantage, demonstrating robustness in this setting.

\subsection{Ablation Studies}

We conduct two complementary ablation studies on LVBench.
The architecture ablation examines the roles of the shared loop, QA-inspector, and layered memory, while the training ablation examines the effects of different algorithms used to optimize the complete VAMR architecture.
Further details are provided in Appendix~\ref{app:ablation-details}.

\begin{figure}[t]
\centering
\includegraphics[width=0.49\textwidth]{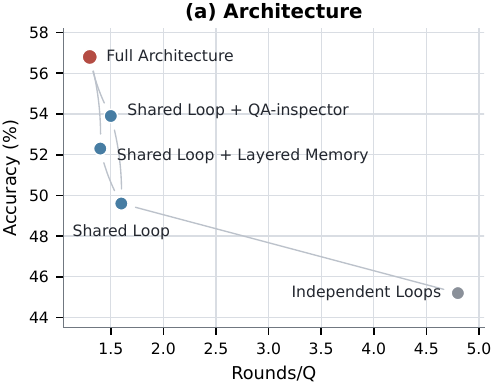}
\hfill
\includegraphics[width=0.49\textwidth]{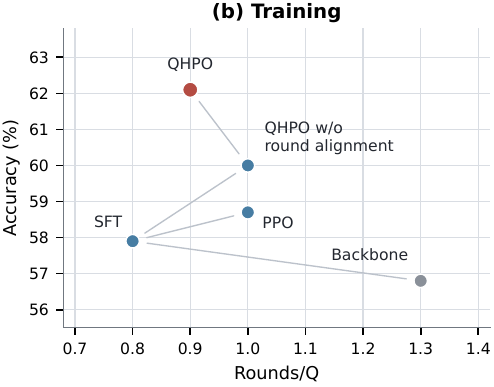}
\caption{Ablation studies on LVBench.
Each point plots overall accuracy against reasoning rounds per question.
Panel (a) isolates the shared loop, QA-inspector, and layered memory using the same Qwen3.5-9B model.
Panel (b) compares the backbone, SFT, PPO, QHPO w/o round alignment, and QHPO.
Lines connect configurations separated by one controlled architectural or training change.
Moving toward the upper left indicates higher accuracy with fewer reasoning rounds.}
\label{fig:ablation-results}
\end{figure}

\paragraph{Architecture ablation.}
This ablation examines the contributions of the shared loop, QA-inspector, and layered memory.
All configurations use the same Qwen3.5-9B model.
Starting from independent loops, we introduce the shared loop with vanilla perception and memory, then replace either vanilla component with the QA-inspector or layered memory before combining both in the full architecture.
As shown in Figure~\ref{fig:ablation-results}(a), replacing independent loops with this vanilla shared loop increases accuracy from 45.2 to 49.6 while reducing Rounds/Q from 4.8 to 1.6 and Frames/Q from 210.7 to 98.7.
The shared topology therefore provides the primary efficiency gain and also improves accuracy without specialized perception or memory.
Replacing the corresponding vanilla component with the QA-inspector or layered memory further raises accuracy to 53.9 and 52.3, respectively, while reducing visual workload.
The larger gain from the QA-inspector identifies question-conditioned perception as an important source of architectural accuracy, while layered memory improves evidence organization within the shared state.
Combining them in the full VAMR architecture further increases accuracy to 56.8 with only 1.3 Rounds/Q, improving over independent loops by 11.6 points while reducing reasoning rounds by 72.9\%.
These results show that the three components are complementary: the shared loop enables coordinated execution, while the QA-inspector and layered memory improve targeted clue retrieval and evidence organization within that loop.

\paragraph{Training ablation.}
All training variants use the complete VAMR architecture, ensuring that the comparison isolates the effect of policy optimization.
We begin with the pretrained backbone, apply SFT to learn the shared interaction protocol, and then compare standard PPO with QHPO without and with round alignment.
Both QHPO variants use question-level values, while round alignment assigns each question advantage only to rounds whose actions serve that question.
All reinforcement learning variants start from the same SFT checkpoint and use the same data and optimization settings.
As shown in Figure~\ref{fig:ablation-results}(b), SFT raises accuracy from 56.8 to 57.9 while reducing Rounds/Q from 1.3 to 0.8, improving both accuracy and sequential-round efficiency.
Standard PPO reaches 58.7 accuracy, QHPO without round alignment reaches 60.0, and full QHPO reaches 62.1.
Full QHPO therefore outperforms PPO by 3.4 points under the same architecture, data, and optimization budget.
Question-level values already improve accuracy without alignment, while aligning each question advantage with its directly served rounds provides a further 2.1-point gain.

\subsection{Analysis}

We conduct two controlled analyses to characterize how the measured gains change with question density and evidence overlap.
Further details are provided in Appendix~\ref{app:analysis-details}.

\begin{figure}[t]
\centering
\includegraphics[width=\textwidth]{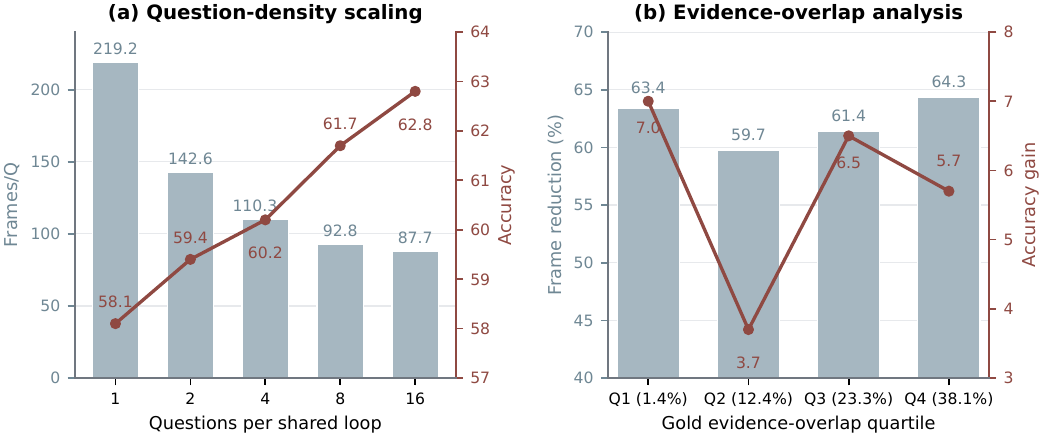}
\caption{Analysis of cross-question sharing on LVBench. Panel (a) reports accuracy and Frames/Q as more questions share one loop under controlled input information. Panel (b) reports visual-frame reduction and accuracy gain over independent loops across relative evidence-overlap quartiles. Mean gold evidence pair overlap is shown in parentheses.}
\label{fig:analysis-results}
\end{figure}

\paragraph{Question-density scaling.}
This analysis examines how VAMR scales as more questions from the same video are solved within one shared loop.
We select all 48 LVBench videos containing at least 16 questions and randomly sample 16 questions for each video, yielding a fixed evaluation set of 768 questions.
We then vary the number of questions solved in each shared loop with $|Q|\in\{1,2,4,8,16\}$ while keeping the videos, questions, input information, and inference settings unchanged.
As shown in Figure~\ref{fig:analysis-results}(a), increasing question density from 1 to 16 raises accuracy from 58.1 to 62.8 while reducing Frames/Q from 219.2 to 87.7.
As question density increases, the shared loop must coordinate a larger and more diverse set of evidence needs, making multi-question reasoning increasingly challenging.
Nevertheless, VAMR achieves higher accuracy while processing fewer frames per question, demonstrating clear advantages and robust generalization as question density increases and fulfilling its design objective for high-density multi-question video understanding.

\paragraph{Evidence overlap analysis.}
We compare VAMR trained through SFT and QHPO with an independent-loop counterpart trained through SFT and PPO.
This analysis examines whether VAMR's gains depend on different questions requiring evidence from the same temporal regions.
For each video, gold evidence pair overlap measures the proportion of question pairs whose annotated evidence intervals have a positive-length temporal intersection.
We rank all 103 LVBench videos by this score and divide them into four equal-sized quartiles from lowest to highest overlap.
Figure~\ref{fig:analysis-results}(b) shows that VAMR improves accuracy in every quartile, while frame reduction remains substantial without increasing monotonically with evidence overlap.
Notably, Q1 has only 1.4\% pair overlap, yet VAMR improves accuracy by 7.0 points and reduces frames by 63.4\%.
The strong Q1 result rules out high evidence overlap as a prerequisite for VAMR and is consistent with the benefit of progressively refining a shared video understanding across diverse evidence needs.

\section{Conclusion}

We presented \method{}, which coordinates questions about one video within a shared agent trajectory.
Its policy invokes tools and resolves questions at different times, supported by question-conditioned perception and layered memory that preserve reusable video understanding and question-specific evidence.
\qhpo{} optimizes this process through question-level value estimation and direct round alignment.
Across three benchmarks, VAMR achieves the highest accuracy and fewest reasoning rounds under our unified evaluation setting.
On LVBench, it outperforms VideoARM by 4.3 points while reducing reasoning rounds and processed frames by 85.9\% and 61.4\%.
Ablations attribute the main efficiency gain to the shared loop, with question-conditioned perception, layered memory, and QHPO further improving accuracy.
Overall, shared video reasoning reduces repeated interaction without sacrificing answer quality across diverse long-video settings.

\clearpage
\section*{AI Use Statement}
Generative AI tools were used to assist with language editing and code development. All AI assisted material was reviewed and verified by the authors, who take full responsibility for the final
manuscript and reported results.

\section*{Reproducibility Statement}

Sections~\ref{sec:method} and~\ref{sec:training} specify the shared-loop formulation, VAMR architecture, and QHPO objective.
The appendices from Appendix~\ref{app:agentic-framework-details} through
Appendix~\ref{app:analysis-details} document the execution protocol, SFT
construction, optimization hyperparameters, dataset partitions, baseline
implementations, metric accounting, and the settings and numerical results for
the ablation and analysis experiments.

\bibliographystyle{iclr2027_conference}
\bibliography{references}

@inproceedings{yao2023react,
  title     = {{ReAct}: Synergizing Reasoning and Acting in Language Models},
  author    = {Yao, Shunyu and Zhao, Jeffrey and Yu, Dian and Du, Nan and
               Shafran, Izhak and Narasimhan, Karthik and Cao, Yuan},
  booktitle = {International Conference on Learning Representations},
  year      = {2023}
}

@article{schulman2017proximal,
  title   = {Proximal Policy Optimization Algorithms},
  author  = {Schulman, John and Wolski, Filip and Dhariwal, Prafulla and
             Radford, Alec and Klimov, Oleg},
  journal = {arXiv preprint arXiv:1707.06347},
  year    = {2017}
}

@inproceedings{schulman2016gae,
  title     = {High-Dimensional Continuous Control Using Generalized Advantage
               Estimation},
  author    = {Schulman, John and Moritz, Philipp and Levine, Sergey and
               Jordan, Michael I. and Abbeel, Pieter},
  booktitle = {International Conference on Learning Representations},
  year      = {2016}
}

@inproceedings{lei2020multiquestion,
  title     = {Multi-Question Learning for Visual Question Answering},
  author    = {Lei, Chenyi and Wu, Lei and Liu, Dong and Li, Zhao and Wang,
               Guoxin and Tang, Haihong and Li, Houqiang},
  booktitle = {AAAI Conference on Artificial Intelligence},
  pages     = {11328--11335},
  year      = {2020}
}

@article{pei2025collective,
  title   = {Guiding Audio-Visual Question Answering with Collective Question
             Reasoning},
  author  = {Pei, Baoqi and Huang, Yifei and Chen, Guo and Xu, Jilan and Wang,
             Yali and Wang, Limin and Lu, Tong and Qiao, Yu and Wu, Fei},
  journal = {International Journal of Computer Vision},
  volume  = {133},
  number  = {10},
  pages   = {6912--6929},
  year    = {2025}
}

@article{liang2025isolated,
  title   = {Beyond Isolated Facts: Synthesizing Narrative and Grounded
             Supervision for {VideoQA}},
  author  = {Liang, Jianxin and Yue, Tan and Wang, Yuxuan and Wang, Yueqian and
             Yin, Zhihan and Zhang, Huishuai and Zhao, Dongyan},
  journal = {arXiv preprint arXiv:2509.24445},
  year    = {2025}
}

@inproceedings{chen2025cgbench,
  title     = {{CG-Bench}: Clue-Grounded Question Answering Benchmark for Long
               Video Understanding},
  author    = {Chen, Guo and Liu, Yicheng and Huang, Yifei and He, Yuping and
               Pei, Baoqi and Xu, Jilan and Wang, Yali and Lu, Tong and Wang,
               Limin},
  booktitle = {International Conference on Learning Representations},
  year      = {2025}
}

@inproceedings{xiao2021nextqa,
  title     = {{NExT-QA}: Next Phase of Question-Answering to Explaining
               Temporal Actions},
  author    = {Xiao, Junbin and Shang, Xindi and Yao, Angela and Chua,
               Tat-Seng},
  booktitle = {IEEE/CVF Conference on Computer Vision and Pattern Recognition},
  year      = {2021}
}

@inproceedings{li2023intentqa,
  title     = {{IntentQA}: Context-Aware Video Intent Reasoning},
  author    = {Li, Jiapeng and Wei, Ping and Han, Wenjuan and Fan, Lifeng},
  booktitle = {IEEE/CVF International Conference on Computer Vision},
  year      = {2023}
}

@inproceedings{mangalam2023egoschema,
  title     = {{EgoSchema}: A Diagnostic Benchmark for Very Long-Form Video
               Language Understanding},
  author    = {Mangalam, Karttikeya and Akshulakov, Raiymbek and Malik,
               Jitendra},
  booktitle = {Advances in Neural Information Processing Systems},
  year      = {2023}
}

@inproceedings{fu2025videomme,
  title     = {{Video-MME}: {The First-Ever Comprehensive Evaluation Benchmark
               of Multi-Modal LLMs in Video Analysis}},
  author    = {Fu, Chaoyou and Dai, Yuhan and Luo, Yongdong and Li, Lei and
               Ren, Shuhuai and Zhang, Renrui and Wang, Zihan and Zhou, Chenyu
               and Shen, Yunhang and Zhang, Mengdan and others},
  booktitle = {IEEE/CVF Conference on Computer Vision and Pattern Recognition},
  year      = {2025}
}

@inproceedings{yan2026ava,
  title     = {{AVA}: Towards Agentic Video Analytics with Vision Language
               Models},
  author    = {Yan, Yuxuan and Jiang, Shiqi and Cao, Ting and Yang, Yifan and
               Yang, Qianqian and Shu, Yuanchao and Yang, Yuqing and Qiu,
               Lili},
  booktitle = {USENIX Symposium on Networked Systems Design and Implementation},
  year      = {2026}
}

@article{aldahoul2024moderation,
  title   = {Advancing Content Moderation: Evaluating Large Language Models for
             Detecting Sensitive Content Across Text, Images, and Videos},
  author  = {AlDahoul, Nouar and Tan, Myles Joshua Toledo and Kasireddy,
             Harishwar Reddy and Zaki, Yasir},
  journal = {arXiv preprint arXiv:2411.17123},
  year    = {2024}
}

@inproceedings{wang2025lvbench,
  title     = {{LVBench}: An Extreme Long Video Understanding Benchmark},
  author = {Wang, Weihan and He, Zehai and Hong, Wenyi and Cheng, Yean and
          Zhang, Xiaohan and Qi, Ji and Gu, Xiaotao and Huang, Shiyu and
          Xu, Bin and Dong, Yuxiao and Ding, Ming and Tang, Jie},
  booktitle = {IEEE/CVF International Conference on Computer Vision},
  year      = {2025}
}

@inproceedings{wu2024longvideobench,
  title     = {{LongVideoBench}: A Benchmark for Long-Context Interleaved
               Video-Language Understanding},
  author    = {Wu, Haoning and Li, Dongxu and Chen, Bei and Li, Junnan},
  booktitle = {Advances in Neural Information Processing Systems},
  year      = {2024}
}

@inproceedings{ye2025tstar,
  title     = {Re-Thinking Temporal Search for Long-Form Video Understanding},
  author    = {Ye, Jinhui and Wang, Zihan and Sun, Haosen and Chandrasegaran,
               Keshigeyan and Durante, Zane and Eyzaguirre, Cristobal and Bisk,
               Yonatan and Niebles, Juan Carlos and Adeli, Ehsan and Fei-Fei,
               Li and Wu, Jiajun and Li, Manling},
  booktitle = {IEEE/CVF Conference on Computer Vision and Pattern Recognition},
  year      = {2025}
}

@article{cheng2025videoholmes,
  title   = {{Video-Holmes}: Can {MLLM} Think Like Holmes for Complex Video
             Reasoning?},
  author  = {Cheng, Junhao and Ge, Yuying and Wang, Teng and Ge, Yixiao and
             Liao, Jing and Shan, Ying},
  journal = {arXiv preprint arXiv:2505.21374},
  year    = {2025}
}

@misc{qwen2026qwen35,
  title        = {{Qwen3.5-9B}},
  author       = {{Qwen Team}},
  year         = {2026},
  howpublished = {Hugging Face model card},
  url          = {https://huggingface.co/Qwen/Qwen3.5-9B}
}

@misc{alibaba2026qwen37plus,
  author       = {{Alibaba Cloud}},
  title        = {{Qwen3.7-Plus}},
  year         = {2026},
  howpublished = {Alibaba Cloud Model Studio documentation},
  url          = {https://help.aliyun.com/zh/model-studio/qwen3-7-plus}
}

@misc{alibaba2026qwen3asr,
  author       = {{Alibaba Cloud}},
  title        = {{Qwen3-ASR-Flash}},
  year         = {2026},
  howpublished = {Alibaba Cloud Model Studio documentation},
  url          = {https://help.aliyun.com/zh/model-studio/asr-model}
}

@inproceedings{feng2025videor1,
  title   = {{Video-R1}: Reinforcing Video Reasoning in {MLLMs}},
  author  = {Feng, Kaituo and Gong, Kaixiong and Li, Bohao and Guo, Zonghao and
             Wang, Yibing and Peng, Tianshuo and Wu, Junfei and Zhang, Xiaoying
             and Wang, Benyou and Yue, Xiangyu},
  booktitle = {Advances in Neural Information Processing Systems},
  year    = {2025}
}

@inproceedings{park2025deepvideor1,
  title   = {{DeepVideo-R1}: Video Reinforcement Fine-Tuning via
             Difficulty-Aware Regressive {GRPO}},
  author  = {Park, Jinyoung and Na, Jeehye and Kim, Jinyoung and Kim, Hyunwoo J.},
  booktitle = {Advances in Neural Information Processing Systems},
  year    = {2025}
}

@inproceedings{wang2024videotree,
  title     = {{VideoTree}: Adaptive Tree-Based Video Representation for {LLM}
               Reasoning on Long Videos},
  author    = {Wang, Ziyang and Yu, Shoubin and Stengel-Eskin, Elias and Yoon,
               Jaehong and Cheng, Feng and Bertasius, Gedas and Bansal, Mohit},
  booktitle = {IEEE/CVF Conference on Computer Vision and Pattern Recognition},
  year      = {2025},
  note      = {arXiv:2405.19209}
}

@article{he2025framethinker,
  title   = {{FrameThinker}: Learning to Think with Long Videos via Multi-Turn
             Frame Spotlighting},
  author  = {He, Zefeng and Qu, Xiaoye and Li, Yafu and Huang, Siyuan and Liu,
             Daizong and Cheng, Yu},
  journal = {arXiv preprint arXiv:2509.24304},
  year    = {2025}
}

@inproceedings{pan2025timesearchr,
  title     = {{TimeSearch-R}: Adaptive Temporal Search for Long-Form Video
               Understanding via Self-Verification Reinforcement Learning},
  author    = {Pan, Junwen and Zhang, Qizhe and Zhang, Rui and Lu, Ming and Wan,
               Xin and Zhang, Yuan and Liu, Chang and She, Qi},
  booktitle = {International Conference on Learning Representations},
  year      = {2026}
}

@inproceedings{wang2025videorts,
  title     = {{Video-RTS}: Rethinking Reinforcement Learning and Test-Time
               Scaling for Efficient and Enhanced Video Reasoning},
  author    = {Wang, Ziyang and Yoon, Jaehong and Yu, Shoubin and Islam, Md
               Mohaiminul and Bertasius, Gedas and Bansal, Mohit},
  booktitle = {Conference on Empirical Methods in Natural Language Processing},
  year      = {2025}
}

@inproceedings{zhang2025dvd,
  title     = {Deep Video Discovery: Agentic Search with Tool Use for Long-Form
               Video Understanding},
  author    = {Zhang, Xiaoyi and Jia, Zhaoyang and Guo, Zongyu and Li, Jiahao
               and Li, Bin and Li, Houqiang and Lu, Yan},
  booktitle = {Advances in Neural Information Processing Systems},
  year      = {2025}
}

@inproceedings{yeo2026worldmm,
  title     = {{WorldMM}: Dynamic Multimodal Memory Agent for Long Video
               Reasoning},
  author    = {Yeo, Woongyeong and Kim, Kangsan and Yoon, Jaehong and Hwang,
               Sung Ju},
  booktitle = {IEEE/CVF Conference on Computer Vision and Pattern Recognition},
  year      = {2026}
}

@inproceedings{yang2026longvt,
  title     = {{LongVT}: Incentivizing ``Thinking with Long Videos'' via Native
               Tool Calling},
  author    = {Yang, Zuhao and Wang, Sudong and Zhang, Kaichen and Wu, Keming
               and Leng, Sicong and Zhang, Yifan and Li, Bo and Qin, Chengwei
               and Lu, Shijian and Li, Xingxuan and Bing, Lidong},
  booktitle = {IEEE/CVF Conference on Computer Vision and Pattern Recognition},
  year      = {2026}
}

@inproceedings{zeng2026videoo3,
  title     = {{Video-o3}: Native Interleaved Clue Seeking for Long Video
               Multi-Hop Reasoning},
  author    = {Zeng, Xiangyu and Zhang, Zhiqiu and Zhu, Yuhan and Li, Xinhao and
               Wang, Zikang and Ma, Changlian and Zhang, Qingyu and Huang,
               Zizheng and Ouyang, Kun and Jiang, Tianxiang and Yan, Ziang and
               Wang, Yi and Zhang, Hongjie and Wang, Yali and Wang, Limin},
  booktitle = {International Conference on Machine Learning},
  year      = {2026}
}

@article{yang2026paravt,
  title   = {{ParaVT}: Taming the Tool Prior Paradox for Parallel Tool Use in
             Agentic Video Reinforcement Learning},
  author = {Yang, Zuhao and Zhang, Kaichen and Wang, Sudong and Wu, Keming and
          Yang, Zhongyu and Li, Bo and Qi, Xiaojuan and Lu, Shijian and
          Li, Xingxuan and Bing, Lidong},
  journal = {arXiv preprint arXiv:2605.20342},
  year    = {2026}
}

@inproceedings{zhang2026eva,
  title     = {{EVA}: Efficient Reinforcement Learning for End-to-End Video
               Agent},
  author    = {Zhang, Yaolun and Wang, Ruohui and Wang, Jiahao and Tang, Yepeng
               and Zheng, Xuanyu and Duan, Haonan and Lu, Hao and Deng,
               Hanming and Lu, Lewei},
  booktitle = {IEEE/CVF Conference on Computer Vision and Pattern Recognition},
  year      = {2026}
}

@article{pang2025mrvideo,
  title   = {{MR. Video}: ``MapReduce'' Is the Principle for Long Video
             Understanding},
  author  = {Pang, Ziqi and Wang, Yu-Xiong},
  journal = {arXiv preprint arXiv:2504.16082},
  year    = {2025}
}

@article{fu2025lover1,
  title   = {{LOVE-R1}: Advancing Long Video Understanding with an Adaptive
             Zoom-in Mechanism via Multi-Step Reasoning},
  author  = {Fu, Shenghao and Yang, Qize and Li, Yuan-Ming and Wei, Xihan and
             Xie, Xiaohua and Zheng, Wei-Shi},
  journal = {arXiv preprint arXiv:2509.24786},
  year    = {2025}
}

@inproceedings{qiu2026longvideor1,
  title     = {{LongVideo-R1}: Smart Navigation for Low-Cost Long Video
               Understanding},
  author    = {Qiu, Jihao and Xie, Lingxi and Huo, Xinyue and Tian, Qi and Ye,
               Qixiang},
  booktitle = {IEEE/CVF Conference on Computer Vision and Pattern Recognition},
  year      = {2026}
}

@inproceedings{li2026lenswalk,
  title     = {{LensWalk}: Agentic Video Understanding by Planning How You See in
               Videos},
  author    = {Li, Keliang and Li, Yansong and Shen, Hongze and Liu, Mengdi and
               Chang, Hong and Shan, Shiguang},
  booktitle = {IEEE/CVF Conference on Computer Vision and Pattern Recognition},
  year      = {2026}
}

@inproceedings{qiu2026videoseal,
  title     = {{VideoSEAL}: Mitigating Evidence Misalignment in Agentic Long
               Video Understanding by Decoupling Answer Authority},
  author    = {Qiu, Chenhao and Zhang, Yechao and Luo, Xin and Song, Shien and
               Liu, Xusheng},
  booktitle = {International Conference on Machine Learning},
  year      = {2026}
}

@article{wei2025multiturncredit,
  title   = {Reinforcing Multi-Turn Reasoning in {LLM} Agents via Fine-Grained
             Reward Structure and Credit Assignment},
  author  = {Wei, Quan and Zeng, Siliang and Li, Chenliang and Wang, Zhongruo
             and Brown, William and Frunza, Oana and Deng, Wei and Schneider,
             Anderson and Nevmyvaka, Yuriy and Zhao, Yang Katie and Garcia,
             Alfredo and Hong, Mingyi},
  journal = {arXiv preprint arXiv:2505.11821},
  year    = {2025}
}

@article{luo2025agentlightning,
  title   = {Agent Lightning: Train Any {AI} Agents with Reinforcement
             Learning},
  author  = {Luo, Xufang and Zhang, Yuge and He, Zhiyuan and Wang, Zilong and
             Zhao, Siyun and Li, Dongsheng and Qiu, Luna K. and Yang, Yuqing},
  journal = {arXiv preprint arXiv:2508.03680},
  year    = {2025}
}

@article{bai2025qwen3vl,
  title   = {{Qwen3-VL} Technical Report},
  author  = {Bai, Shuai and Cai, Yuxuan and Chen, Ruizhe and Chen, Keqin and
             Chen, Xionghui and Cheng, Zesen and Deng, Lianghao and Ding, Wei
             and Gao, Chang and Ge, Chunjiang and others},
  journal = {arXiv preprint arXiv:2511.21631},
  year    = {2025}
}

@inproceedings{yin2025videoarm,
  title     = {{VideoARM}: Agentic Reasoning over Hierarchical Memory for
               Long-Form Video Understanding},
  author    = {Yin, Yufei and Meng, Qianke and Chen, Minghao and Ding, Jiajun
               and Shao, Zhenwei and Yu, Zhou},
  booktitle = {IEEE/CVF Conference on Computer Vision and Pattern Recognition},
  year      = {2026},
  note      = {arXiv:2512.12360}
}

@inproceedings{wang2026videohv,
  title     = {Think, Then Verify: A Hypothesis-Verification Multi-Agent
               Framework for Long Video Understanding},
  author    = {Wang, Zheng and Chen, Haoran and Qin, Haoxuan and Wei, Zhipeng
               and Qian, Tianwen and Bai, Cong},
  booktitle = {IEEE/CVF Conference on Computer Vision and Pattern Recognition},
  year      = {2026}
}

@inproceedings{jin2025searchr1,
  title     = {{Search-R1}: Training {LLMs} to Reason and Leverage Search Engines
               with Reinforcement Learning},
  author    = {Jin, Bowen and Zeng, Hansi and Yue, Zhenrui and Yoon, Jinsung and
               Arik, Sercan and Wang, Dong and Zamani, Hamed and Han, Jiawei},
  booktitle = {Conference on Language Modeling},
  year      = {2025}
}

@article{li2025torl,
  title   = {{ToRL}: Scaling Tool-Integrated {RL}},
  author  = {Li, Xuefeng and Zou, Haoyang and Liu, Pengfei},
  journal = {arXiv preprint arXiv:2503.23383},
  year    = {2025}
}

@article{wang2025ragen,
  title   = {{RAGEN}: Understanding Self-Evolution in {LLM} Agents via
             Multi-Turn Reinforcement Learning},
  author  = {Wang, Zihan and Wang, Kangrui and Wang, Qineng and Zhang, Pingyue
             and Li, Linjie and Yang, Zhengyuan and Jin, Xing and Yu, Kefan and
             Nguyen, Minh Nhat and Liu, Licheng and others},
  journal = {arXiv preprint arXiv:2504.20073},
  year    = {2025}
}

@inproceedings{zhou2024archer,
  title     = {{ArCHer}: Training Language Model Agents via Hierarchical
               Multi-Turn {RL}},
  author    = {Zhou, Yifei and Zanette, Andrea and Pan, Jiayi and Levine, Sergey
               and Kumar, Aviral},
  booktitle = {International Conference on Machine Learning},
  year      = {2024}
}

@inproceedings{wang2025stepsearch,
  title     = {{StepSearch}: Igniting {LLMs} Search Ability via Step-Wise
               Proximal Policy Optimization},
  author    = {Wang, Ziliang and Zheng, Xuhui and An, Kang and Ouyang, Cijun and
               Cai, Jialu and Wang, Yuhang and Wu, Yichao},
  booktitle = {Conference on Empirical Methods in Natural Language Processing},
  year      = {2025}
}

@article{feng2025gigpo,
  title   = {Group-in-Group Policy Optimization for {LLM} Agent Training},
  author  = {Feng, Lang and Xue, Zhenghai and Liu, Tingcong and An, Bo},
  journal = {arXiv preprint arXiv:2505.10978},
  year    = {2025}
}

@article{liang2025univa,
  title   = {{UniVA}: Universal Video Agent towards Open-Source Next-Generation
             Video Generalist},
  author  = {Liang, Zhengyang and Zhang, Daoan and Zhou, Huichi and Huang, Rui
             and Li, Bobo and Zhang, Yuechen and Wu, Shengqiong and Wang,
             Xiaohan and Luo, Jiebo and Liao, Lizi and Fei, Hao},
  journal = {arXiv preprint arXiv:2511.08521},
  year    = {2025}
}

\appendix
\section{Limitations}
\label{app:limitations}

The primary scope limitation of VAMR is its focus on scenarios in which multiple questions examine different aspects of the same video. This setting is reflected in most widely used video-QA benchmarks, including NExT-QA~\citep{xiao2021nextqa}, IntentQA~\citep{li2023intentqa}, Video-MME~\citep{fu2025videomme}, LVBench~\citep{wang2025lvbench}, and CG-Bench~\citep{chen2025cgbench}, which provide multiple QA pairs for each source video. It also arises in practical applications that assess a recording or stream for different events, objects, actions, or policy violations, such as content moderation and security inspection~\citep{aldahoul2024moderation}. Future work can explore alternative architectures and policy-optimization algorithms for the shared-loop setting introduced here.

\section{More Background}
\label{app:more-background}

\paragraph{Additional video-agent methods.}
Recent learned video reasoners include LOVE-R1~\citep{fu2025lover1}, LongVT~\citep{yang2026longvt}, LongVideo-R1~\citep{qiu2026longvideor1}, EVA~\citep{zhang2026eva}, Video-o3~\citep{zeng2026videoo3}, and ParaVT~\citep{yang2026paravt}.
Complementary systems structure long-video inference through distributed processing, persistent graph or multimodal memory, planned observation, or decoupled planning and perception, including MR. Video~\citep{pang2025mrvideo}, AVA~\citep{yan2026ava}, WorldMM~\citep{yeo2026worldmm}, LensWalk~\citep{li2026lenswalk}, and VideoSEAL~\citep{qiu2026videoseal}.
ParaVT parallelizes tool calls for one query, whereas VAMR coordinates joint actions across multiple asynchronously resolved questions.
MR. Video constructs a question-independent video representation before answering queries separately, whereas VAMR lets unresolved questions jointly direct online evidence acquisition and state updates.
These systems optimize the processing of an individual query.

\paragraph{Credit assignment for tool-using agents.}
Credit assignment for multi-turn agents has likewise been studied through dense turn-level rewards and hierarchical decomposition of agent trajectories~\citep{wei2025multiturncredit,luo2025agentlightning}.
These approaches refine where credit is assigned within a trajectory associated with one task outcome.
QHPO instead separates the outcomes of multiple asynchronously terminating questions in the same trajectory and aligns each question-level advantage with the rounds that directly serve it.

\section{Agentic Framework Details}
\label{app:agentic-framework-details}

\paragraph{Inference configuration.}
The active question window contains five questions.
For a video with question set $Q$, the normal-loop budget is $T_{\max}=\max(12,4|Q|)$ controller rounds.
The episode terminates when this budget is exhausted, and no additional rounds are granted.
Each round permits at most six tool calls.
The policy model uses a 4,096-token output budget and a 40,960-token context window.
The policy emits the explicit \texttt{<think>...</think>} reasoning format specified by the prompt.

\paragraph{Shared loop and question scheduling.}
The policy model operates on the five unresolved questions in the active window, while the remaining questions stay in an ordered backlog and are not exposed to the model.
An accepted answer removes its question from the active window, after which memory is consolidated and the vacant slot is filled from the backlog in input order.
The shared story layer persists across these transitions, giving newly activated questions access to the previously accumulated video context.

\paragraph{Action protocol.}
Each policy-model response contains an \texttt{answers} list and a \texttt{tool\_calls} list.
A turn may submit several answers and issue up to six tool calls, with answers excluded from the call limit.
An answer commit is accepted only when its question is open, the same action contains no duplicate submission for that question, and the submitted option exactly matches one listed candidate.
A tool call must respect the six-call limit, reference an open question for visual retrieval, avoid retrieving and answering the same question in one action, and use parseable timestamps within the video range.
Fine-grained retrieval must stay within a previously inspected interval and span at most 30 seconds, while ASR accepts no question identifiers, options, or frame-count argument and spans at most 90 seconds.
Invalid components are removed without changing other valid actions in the same response, and invalid arguments are not automatically repaired.

\paragraph{Tool interfaces.}
A visual call specifies a temporal interval and target question identifiers, and the harness supplies their question text and answer options to the perception model.
The QA-inspector samples 30, 60, 90, or 150 frames from the requested interval and returns one shared story summary together with per-question clues and option-level evidence.
The fine-grained QA-inspector revisits a previously localized interval of at most 30 seconds and processes high-resolution frames without the sparse mosaic used by the QA-inspector.
ASR accepts only a temporal interval of at most 90 seconds and returns its transcript.

\paragraph{Memory transition.}
The resulting consolidation updates the shared story layer with reusable visual and speech information, routes question-specific evidence to the corresponding clue memories, discards the clue memories of completed questions, and clears the buffer.

\section{SFT Construction Details}
\label{app:sft-data-construction}

We collect one supervised example from each policy-model turn in an executed multi-question trajectory using the two-stage procedure in Eq.~\ref{eq:two-stage-sft}.
Qwen3-VL-32B serves as the teacher model in both stages~\citep{bai2025qwen3vl}.

\paragraph{Privileged action selection.}
The first stage receives the active questions and layered memory together with the correct answers and annotated evidence intervals.
These annotations guide the selection of retrieval windows, shared calls for questions with nearby evidence, and answer submissions already supported by memory.
The correct answer alone is not treated as evidence, so unsupported questions must continue retrieval.
When annotated evidence intervals are unavailable, Stage 1 can use correct answers as the only privileged signal and continue retrieval until the public memory supports answer submission.
Evidence intervals facilitate targeted trajectory construction but are not required by the learned policy or at inference time.
The selected joint action is executed under the same action protocol used at inference, and only accepted submissions and successful tool calls advance the trajectory.

\paragraph{Reasoning without privileged information.}
After fixing the executed action, the second stage removes the correct answers and evidence intervals and generates reasoning from the active questions, pre-action memory, and action alone.
The reasoning must justify answer submissions with evidence already visible in memory and explain the information sought by each tool call.
Samples that rely on hidden annotations or unavailable evidence are discarded.
The final target pairs the resulting public-state reasoning with the fixed joint action.

\section{QHPO Training Details}
\label{app:training-details}

This section provides the implementation details needed to reproduce the QHPO procedure described in Section~\ref{sec:qhpo}.

\paragraph{Rollout and optimization.}
Each training video produces one complete shared-loop trajectory from the current policy model, and every round records the public state, sampled joint action, rollout-policy likelihood, and active questions.
QHPO uses all 584 videos in the CG-Bench policy-optimization partition for three fresh on-policy passes: one critic warm-up pass with the actor frozen, followed by two joint actor and critic passes.
We collect 8 video trajectories per update, optimize 4-video minibatches for three PPO epochs and two critic epochs, and use $\gamma=1$, GAE $\lambda=0.95$, a PPO clip ratio of $0.2$, and a KL coefficient of $0.02$ against the frozen SFT policy.
The actor is a rank-64 LoRA adapter with scale 128 and learning rate $10^{-6}$.
The critic applies LayerNorm and a linear head followed by a sigmoid over
detached policy-model features, with learning rate $10^{-4}$.
After critic initialization, subsequent batches update the actor and critic together before collecting new trajectories from the updated policy.

\paragraph{Question outcome.}
Question $q_i$ receives $R_i=1$ when its submitted answer equals the gold answer and receives zero otherwise.
Gold answers are used only for this post-rollout judgment and never enter the actor or critic input.

\paragraph{Question-level GAE.}
For a question active from round $s_i$ through its resolution round $e_i$, we set $r_{t,i}=0$ for $s_i\le t<e_i$, $r_{e_i,i}=R_i$, and the value after resolution to zero.
We then compute $\delta_{t,i}=r_{t,i}+\gamma V_\phi(B_{t+1},M_{t+1},q_i)-V_\phi(B_t,M_t,q_i)$ and $A_{t,i}=\sum_{k=0}^{e_i-t}(\gamma\lambda)^k\delta_{t+k,i}$ over the rounds in which $q_i$ remains active.

\paragraph{Validated action attribution.}
Malformed components, illegal arguments, invalid question identifiers, and duplicate requests are removed before credit assignment.
An accepted answer submission is attributed to its submitted question, a successful QA-inspector call is attributed to its explicit target questions, and an ASR call is attributed to all questions active in that round.
After this attribution, $A_{t,i}$ is retained only when round $t$ serves $q_i$ and is otherwise set to zero before applying Eq.~\ref{eq:question-horizon-credit}.
If $\mathcal{S}_t=\varnothing$, we set $A_t=0$, so round $t$ contributes no actor update.

\section{Dataset Statistics}
\label{app:datasets}

VAMR is trained and validated exclusively on CG-Bench~\citep{chen2025cgbench}.
We partition its 1,219 videos by video identifier so that questions from the same video never cross splits.
Table~\ref{tab:cgbench-statistics} reports the resulting SFT, \qhpo{}, and validation partitions.

\begin{table}[h]
\caption{Statistics of the CG-Bench training and validation partitions.}
\label{tab:cgbench-statistics}
\begin{center}
\small
\setlength{\tabcolsep}{12pt}
\renewcommand{\arraystretch}{1.15}
\begin{tabular}{lrr}
\toprule
\textbf{Partition} & \textbf{Videos} & \textbf{Question Instances} \\
\midrule
SFT & 610 & 5,775 \\
\qhpo{} & 584 & 6,001 \\
Validation & 25 & 353 \\
\midrule
\textbf{Total} & \textbf{1,219} & \textbf{12,129} \\
\bottomrule
\end{tabular}
\end{center}
\end{table}

Table~\ref{tab:dataset-statistics} compares CG-Bench with the three held-out evaluation splits.
LVBench~\citep{wang2025lvbench} contains hour-long videos with dense groups of questions.
It averages 15.04 questions per video, substantially more than Video-Holmes and LongVideoBench.
This high question density creates greater scope for sharing observations and coordinating evidence acquisition within a single loop.
LongVideoBench~\citep{wu2024longvideobench} covers diverse long-context video understanding tasks, while Video-Holmes~\citep{cheng2025videoholmes} focuses on multi-step reasoning.

\begin{table}[h]
\caption{Video and question statistics for the training and evaluation benchmarks. Average duration is measured over unique videos.}
\label{tab:dataset-statistics}
\begin{center}
\small
\setlength{\tabcolsep}{4pt}
\renewcommand{\arraystretch}{1.15}
\resizebox{0.90\textwidth}{!}{%
\begin{tabular}{lcrrrr}
\toprule
\textbf{Dataset} & \textbf{Split} & \textbf{Videos}
& \textbf{Questions} & \textbf{Questions/Video}
& \textbf{Avg. Duration (min)} \\
\midrule
CG-Bench & Full & 1,219 & 12,129 & 10.0 & 27.7 \\
LVBench & Full & 103 & 1,549 & 15.0 & 68.4 \\
Video-Holmes & Test & 270 & 1,837 & 6.8 & 2.8 \\
LongVideoBench & Validation & 753 & 1,337 & 1.8 & 7.9 \\
\bottomrule
\end{tabular}
}
\end{center}
\end{table}

CG-Bench and LongVideoBench durations are computed from their released metadata, while the LVBench duration follows its official dataset statistics.
The Video-Holmes question file does not provide video duration, so we use the final timestamp in its released segment annotations.

\paragraph{Data-overlap audit.}
We compare the released source video identifiers of CG-Bench against the evaluation annotations and find no overlap with the 103 LVBench videos, 270 Video-Holmes test videos, or 753 LongVideoBench validation videos.

\section{Main Experiment Details}
\label{app:main-experiment-details}

\paragraph{Shared evaluation protocol.}
For each video, every reproduced baseline receives all associated question texts and answer options as read-only reference context, while answers, tool calls, and memory updates remain restricted to the current target question.
LongVideoBench is evaluated on its validation split using raw videos without the provided subtitles.

\paragraph{Statistical consistency.}
Table~\ref{tab:multibench-multirun} reports three inference runs with different seeds for the principal baseline and training comparisons on all three benchmarks.
These deviations quantify run-to-run inference variability on the fixed evaluation sets.

\begin{table}[h]
\caption{Accuracy over three inference runs, reported as mean $\pm$ standard deviation.}
\label{tab:multibench-multirun}
\begin{center}
\footnotesize
\setlength{\tabcolsep}{4pt}
\renewcommand{\arraystretch}{1.1}
\resizebox{0.70\textwidth}{!}{%
\begin{tabular}{lccc}
\toprule
\textbf{Configuration} & \textbf{LVBench} & \textbf{Video-Holmes} & \textbf{LongVideoBench} \\
\midrule
VideoARM & $57.8 \pm 0.5$ & $54.6 \pm 0.4$ & $65.9 \pm 0.4$ \\
Ind.\ Loops (SFT+PPO) & $56.4 \pm 0.4$ & $53.8 \pm 0.3$ & $65.1 \pm 0.3$ \\
SFT & $57.9 \pm 0.3$ & $54.8 \pm 0.2$ & $65.7 \pm 0.3$ \\
PPO & $58.7 \pm 0.5$ & $55.6 \pm 0.4$ & $65.9 \pm 0.5$ \\
QHPO w/o round alignment & $60.0 \pm 0.4$ & $56.4 \pm 0.4$ & $66.1 \pm 0.3$ \\
VAMR / QHPO & $\mathbf{62.1 \pm 0.3}$ & $\mathbf{58.2 \pm 0.3}$ & $\mathbf{66.5 \pm 0.4}$ \\
\bottomrule
\end{tabular}
}
\end{center}
\end{table}

\paragraph{Metric accounting.}
Rounds/Q is the number of valid joint reasoning rounds for a video divided by its number of questions.
A joint round may advance several questions, so this amortized quantity can be below one.
It counts sequential policy-model actions and excludes formatting retries, API retries, and service failures.
End-to-end video reasoning methods have no iterative policy model, so Rounds/Q is reported as N/A.

Frames/Q is the total number of frames processed by visual models for a video divided by its number of questions.
The same frame is counted again whenever it is reprocessed in another round or tool call.
Frames used to construct DVD's database and VideoHV-Agent's frame captions are counted once per video before normalization, whereas all frames processed by VAMR are charged at every visual tool call without deduplication.
Rounds/Q and Frames/Q therefore capture complementary costs: the former measures sequential-round efficiency, while the latter measures visual-frame efficiency.

Model Calls/Q counts model invocations outside the main policy/controller divided by the number of questions.
Table~\ref{tab:agentic-calls} reports this complementary measure for agentic methods.
For DVD and VideoHV-Agent, each nonterminal round issues one external model call, while the final answer round issues none.

\begin{table}[h]
\caption{External model calls per question.}
\label{tab:agentic-calls}
\begin{center}
\footnotesize
\setlength{\tabcolsep}{8pt}
\renewcommand{\arraystretch}{1.1}
\begin{tabular}{lccc}
\toprule
\textbf{Method} & \textbf{LVBench} & \textbf{Video-Holmes} & \textbf{LongVideoBench} \\
\midrule
DVD & 6.3 & 5.3 & 6.5 \\
VideoHV-Agent & 4.7 & 4.3 & 4.4 \\
VideoARM & 2.4 & 2.3 & 2.6 \\
VAMR & 1.2 & 1.5 & 2.3 \\
\bottomrule
\end{tabular}
\end{center}
\end{table}

\paragraph{Direct controls.}
Qwen3.5-9B (Direct) and Qwen3.7-Plus (Direct) independently answer each question from the same 32 uniformly sampled frames.
Neither control invokes external tools or performs iterative retrieval, giving 32 Frames/Q and N/A Rounds/Q on every benchmark.

\paragraph{Qwen3.5-9B + Tools.}
The tool-augmented control uses the pretrained Qwen3.5-9B backbone without VAMR-specific SFT or policy optimization.
It uses the same visual and speech modules as VAMR but runs an independent loop with private memory for each question.
Observations cannot be reused across questions, so all reasoning rounds and visual frames are charged to the corresponding question.

\paragraph{Ind. Loops (SFT+PPO).}
This counterpart uses the same Qwen3.5-9B initialization, QA-inspector, perception models, CG-Bench partitions, Qwen3-VL-32B teacher, and overall SFT and reinforcement-learning budgets as VAMR.
Each trajectory contains one active question with private memory, so observations and policy state are never shared across questions.
Its SFT data follow the same two-stage construction as VAMR: privileged answers and evidence intervals first determine an independent-loop action, after which they are removed before the teacher generates the corresponding reasoning trace.
The resulting SFT policy is further optimized with standard PPO on single-question trajectories.

\paragraph{End-to-end video reasoning.}
Video-R1~\citep{feng2025videor1} and DeepVideo-R1~\citep{park2025deepvideor1} process 32 uniformly sampled frames for each question without an external tool loop.
Their Frames/Q is therefore 32, while Rounds/Q is N/A because neither method performs iterative policy-model actions.

\paragraph{FrameThinker.}
FrameThinker~\citep{he2025framethinker} first reads a small set of overview frames and then selects additional temporal locations as reasoning proceeds.
Frames/Q accumulates the complete visual context received at every reasoning round, including previously observed frames that are processed again.

\paragraph{TimeSearch-R.}
TimeSearch-R~\citep{pan2025timesearchr} constructs an initial preview at 2 FPS, capped at 768 frames, and then performs dynamic temporal search.
The preview and all newly selected frames are counted whenever they reenter the model context, rather than counting only unique frames.

\paragraph{DVD.}
DVD~\citep{zhang2025dvd} captions and indexes the video at 2 FPS before answering questions, and then performs question-specific browsing and inspection over the resulting database.
This video-level cost is divided by the number of questions associated with the video.
Frames used by subsequent question-specific browsing and inspection are then added.
The LVBench evaluation uses the released caption database.

\paragraph{VideoHV-Agent.}
VideoHV-Agent~\citep{wang2026videohv} performs hypothesis generation, distinction, verification, and answer generation after captioning the complete video at 1 FPS.
These captions are constructed once and shared by all questions, so their frame cost is divided by the number of questions associated with the video.
Frames used for detailed verification are counted separately for each question.
The reported Frames/Q includes both components.

\paragraph{VideoARM.}
VideoARM~\citep{yin2025videoarm} dynamically invokes coarse visual sampling, local clip analysis, and speech transcription within an observe, think, act, and memorize loop for each question.
Because it does not construct a question-independent visual database, none of its visual processing is amortized across questions.

\paragraph{Native published results.}
Table~\ref{tab:native-published-results} provides a complementary comparison under the native model configurations of different methods.
DVD obtains 74.2 on LVBench and 71.6 on LongVideoBench, while VideoARM reaches 79.7 and 73.7, respectively.
VAMR achieves 80.4 on LVBench, 62.5 on Video-Holmes, and 72.9 on LongVideoBench.
These results show that VAMR remains highly effective with a strong native model stack.
DVD and VideoARM are representative agentic baselines in the main comparison.

\begin{table}[h]
\caption{Accuracy under native model configurations. Prior-method values are published results, whereas VAMR is evaluated with the model stack shown.}
\label{tab:native-published-results}
\begin{center}
\footnotesize
\setlength{\tabcolsep}{4pt}
\renewcommand{\arraystretch}{1.15}
\resizebox{\textwidth}{!}{%
\begin{tabular}{llccc}
\toprule
\textbf{Method} & \textbf{Backbone / Native Config.} & \textbf{LVBench} & \textbf{Video-Holmes} & \textbf{LongVideoBench} \\
\midrule
LOVE-R1 & Qwen2.5-VL-7B & 48.2 & N/A & 60.1 \\
LongVT & Qwen2.5-VL-7B & 41.3 & N/A & N/A \\
MR. Video & Gemini-2.0-Flash + GPT-4o & 60.8 & N/A & 61.6$^\dagger$ \\
AVA & Qwen2.5-VL-7B + Qwen2.5-32B + Gemini-1.5-Pro & 62.3 & N/A & N/A \\
LongVideo-R1 & Qwen3-8B + Qwen2.5-VL-72B & 50.0 & N/A & N/A \\
EVA & Qwen2.5-VL-7B & 43.3 & 37.2 & 55.0 \\
WorldMM & GPT-5 + GPT-5-mini & 61.9 & N/A & N/A \\
Video-o3 & Qwen2.5-VL-7B & 47.6 & 46.5 & 60.5 \\
ParaVT & Qwen3-VL-8B & 39.8 & N/A & 60.4 \\
LensWalk & OpenAI o3 + GPT-4.1 & 66.8 & N/A & 69.9$^\dagger$ \\
VideoSEAL & Qwen3-8B + Qwen2.5-VL-7B & 55.1 & N/A & 62.0 \\
DVD & OpenAI o3 + GPT-4.1/4.1-mini & 74.2 & N/A & 71.6 \\
VideoARM & OpenAI o3 + GPT-4.1 + whisper-1 & 79.7 & N/A & 73.7 \\
VAMR (training-free) & OpenAI o3 + GPT-4.1 + whisper-1 & 80.4 & 62.5 & 72.9 \\
\bottomrule
\end{tabular}
}
\end{center}
\vspace{2pt}
\footnotesize{N/A denotes an unreported result. $^\dagger$ indicates evaluation
on the LongVideoBench long subset. Native configurations differ in backbones,
frames, subtitles, tools, and inference budgets.}
\end{table}

\section{Ablation Experiment Details}
\label{app:ablation-details}

Tables~\ref{tab:ablation-numeric} and \ref{tab:training-ablation-numeric} report the numerical results for the architecture and training ablations in Section~\ref{sec:experiments}.

\paragraph{Architecture ablation setup.}
All architecture variants use the same Qwen3.5-9B model.
They also use identical LVBench questions, perception models, decoding settings, answer parsers, and interaction budgets.
Independent loops reset policy-model context and memory for every question and provide the no-sharing reference.
Within the shared topology, the vanilla configuration uses a generic visual inspector and a single-layer memory.
The next two configurations replace one vanilla component at a time with the QA-inspector or layered memory, and VAMR combines both replacements.
The vanilla inspector uses the same call interface, frame sampling, execution path, and output schema as the QA-inspector but masks question text and answer options from the retrieval prompt.
The vanilla memory pools shared context and question-specific clues into one memory layer while retaining the same update schedule.
The independent-loop endpoint matches the Qwen3.5-9B + Tools control in Table~\ref{tab:main-results}, while the full-architecture endpoint matches the Backbone row of the training ablation.

\begin{table}[h]
\caption{Numerical results for the architecture ablation on LVBench. Figure~\ref{fig:ablation-results}(a) visualizes these results. All variants use the same Qwen3.5-9B model.}
\label{tab:ablation-numeric}
\begin{center}
\footnotesize
\setlength{\tabcolsep}{6pt}
\renewcommand{\arraystretch}{1.15}
\begin{tabular}{lrrr}
\toprule
\textbf{Configuration} & \textbf{Acc.}
& \textbf{Rounds/Q} & \textbf{Frames/Q} \\
\midrule
Independent loops & 45.2 & 4.8 & 210.7 \\
Shared loop & 49.6 & 1.6 & 98.7 \\
Shared loop + QA-inspector & 53.9 & 1.5 & 82.5 \\
Shared loop + layered memory & 52.3 & 1.4 & 88.9 \\
Full \method{} architecture & 56.8 & 1.3 & 72.4 \\
\bottomrule
\end{tabular}
\end{center}
\end{table}

\paragraph{Training ablation setup.}
All training variants use the complete VAMR architecture and the same evaluation protocol.
PPO and QHPO start from the same SFT checkpoint and share the CG-Bench reinforcement learning partition, rollout budget, decoding settings, and optimization hyperparameters.
PPO uses one round-level value for the shared trajectory.
Both QHPO variants instead estimate a separate advantage $A_{t,i}$ for each active question $q_i$.
We define round alignment by whether the validated joint action at round $t$ serves $q_i$, such as through a targeted visual call or the submission of its answer.
With round alignment, $A_{t,i}$ is retained only at these aligned rounds and is zeroed at all other rounds.
Without round alignment, the same advantage is assigned to every round from the activation of $q_i$ until its resolution, including rounds whose actions serve other questions.

\begin{table}[h]
\caption{Numerical results for the training ablation on LVBench. Figure~\ref{fig:ablation-results}(b) visualizes these results. All variants use the VAMR architecture.}
\label{tab:training-ablation-numeric}
\begin{center}
\footnotesize
\setlength{\tabcolsep}{6pt}
\renewcommand{\arraystretch}{1.15}
\begin{tabular}{lrrr}
\toprule
\textbf{Training Objective} & \textbf{Acc.}
& \textbf{Rounds/Q} & \textbf{Frames/Q} \\
\midrule
Backbone & 56.8 & 1.3 & 72.4 \\
SFT & 57.9 & 0.8 & 76.1 \\
PPO & 58.7 & 1.0 & 82.6 \\
\qhpo{} w/o round alignment & 60.0 & 1.0 & 93.7 \\
\qhpo{} & 62.1 & 0.9 & 84.9 \\
\bottomrule
\end{tabular}
\end{center}
\end{table}

\section{Analysis Experiment Details}
\label{app:analysis-details}

\paragraph{Question-density scaling.}
We select 48 LVBench videos that contain at least 16 questions.
For each video, we use a fixed random seed to sample 16 questions and randomly permute them, yielding 768 questions in total.
These questions are evaluated at every density.
For this controlled analysis only, the policy prompt exposes all 16 question texts and answer options as read-only context at every group size.
At group size $|Q|$, all questions in the group share one loop and persistent
memory, but at most five occupy the active window $B_t$ at a time. The
remainder enter from the backlog as slots become available.
Only questions currently in $B_t$ may be answered or targeted by tools.
This design controls for any accuracy benefit from merely exposing additional test questions and answer options, since changing the group size changes which questions share tool observations and memory without changing the textual information available to the controller.
For a group size of 1, each video is divided into 16 single-question loops.
Groups of 2, 4, 8, and 16 questions are formed from consecutive blocks of this random permutation rather than from temporal or identifier-based neighborhoods, producing 768, 384, 192, 96, and 48 loops, respectively.
All group sizes use the final VAMR model with the same shared-loop implementation, QA-inspector, layered memory, and inference configuration, including the $|Q|=1$ condition.
Memory is reset between groups.

\begin{table}[h]
\caption{Numerical results for the question-density scaling analysis in Figure~\ref{fig:analysis-results}(a). Every setting evaluates the same 768 questions and exposes the same 16 question-option pairs per video to the controller.}
\label{tab:density-numeric}
\begin{center}
\footnotesize
\setlength{\tabcolsep}{9pt}
\renewcommand{\arraystretch}{1.15}
\begin{tabular}{lrrrr}
\toprule
\textbf{Questions/loop}
& \textbf{Loops}
& \textbf{Accuracy}
& \textbf{Rounds/Q}
& \textbf{Frames/Q} \\
\midrule
1  & 768 & 58.1 & 4.4 & 219.2 \\
2  & 384 & 59.4 & 2.5 & 142.6 \\
4  & 192 & 60.2 & 1.6 & 110.3 \\
8  & 96  & 61.7 & 1.1 & 92.8 \\
16 & 48  & 62.8 & 0.9 & 87.7 \\
\bottomrule
\end{tabular}
\end{center}
\end{table}

\begin{table}[h]
\caption{Numerical results for the evidence-overlap analysis in Figure~\ref{fig:analysis-results}(b). Videos are ranked by gold evidence pair overlap and divided into relative quartiles from lowest (Q1) to highest (Q4). Ind. denotes the independent-loop policy trained through SFT followed by standard PPO, while VAMR is trained through SFT followed by QHPO.}
\label{tab:overlap-numeric}
\begin{center}
\footnotesize
\setlength{\tabcolsep}{4pt}
\renewcommand{\arraystretch}{1.15}
\resizebox{\textwidth}{!}{%
\begin{tabular}{lrrrrrrrr}
\toprule
\textbf{Quartile}
& \textbf{Videos}
& \textbf{Pair overlap (\%)}
& \textbf{Ind. Acc.}
& \textbf{VAMR Acc.}
& \textbf{$\Delta$Acc.}
& \textbf{Ind. Frames/Q}
& \textbf{VAMR Frames/Q}
& \textbf{Reduction (\%)} \\
\midrule
Q1 (lowest)  & 26 & 1.4  & 54.0 & 61.0 & +7.0  & 228.4 & 83.5 & 63.4 \\
Q2           & 26 & 12.4 & 52.4 & 56.1 & +3.7  & 226.0 & 91.0 & 59.7 \\
Q3           & 26 & 23.3 & 60.5 & 67.0 & +6.5  & 221.7 & 85.5 & 61.4 \\
Q4 (highest) & 25 & 38.1 & 59.2 & 64.9 & +5.7  & 219.6 & 78.5 & 64.3 \\
\midrule
\textbf{Overall}
& \textbf{103}
& \textbf{18.6}
& \textbf{56.4}
& \textbf{62.1}
& \textbf{+5.7}
& \textbf{224.0}
& \textbf{84.9}
& \textbf{62.1} \\
\bottomrule
\end{tabular}
}
\end{center}
\end{table}

\paragraph{Evidence overlap analysis.}
We compare the independent-loop policy trained through SFT followed by standard PPO with VAMR trained through SFT followed by QHPO on the same 103 LVBench videos and 1,549 questions.
For question $q_i$, let $E_i$ denote the union of its annotated gold evidence intervals.
For a video $V$ with $m$ questions that have valid evidence intervals, we define
\begin{equation}
\operatorname{GoldPairOverlap}(V)
=
\frac{1}{\binom{m}{2}}
\sum_{i<j}
\mathbb{I}\!\left[|E_i\cap E_j|>0\right].
\label{eq:gold-evidence-pair-overlap}
\end{equation}
Of the 1,549 LVBench questions, 1,547 have valid evidence intervals. The other
two are excluded only when computing gold evidence overlap.
Reversed endpoints and timestamp formats are normalized before interval intersections are computed.
Accuracy and Frames/Q are evaluated on all questions, with Frames/Q counting the complete visual-tool workload and repeated processing in separate calls.
Videos are ranked by gold evidence pair overlap and divided into relative quartiles.

\end{document}